%% file: samplepaper.tex
\documentclass[runningheads]{llncs}
\usepackage[T1]{fontenc}
\usepackage{graphicx}
\usepackage{caption}
\usepackage{subcaption}
\usepackage{multirow}

\begin{document}
\title{Exploring Graph Neural Networks for Indian Legal Judgment Prediction}
%
%
\author{Mann Khatri\inst{1}\orcidID{0000-0002-5132-9223} \and
 Mirza Yusuf\inst{1}\orcidID{0000-0002-8293-5381} \and
Rajiv Ratn Shah\inst{1}\orcidID{0000-0003-1028-9373} \and
Ponnurangam Kumaraguru\inst{2}\orcidID{0000-0001-5082-2078}
}
%
\authorrunning{Mann et al.}
%
\institute{Indraprastha Institute of Information Technology, Delhi\\
\email{\{mannk,rajivratn\}@iiitd.ac.in, mirzayusuf1000@gmail.com} \and
International Institute of Information Technology Hyderabad\\
\email{pk.guru@iiit.ac.in}\\
}
\maketitle              
\begin{abstract}
The burdensome impact of a skewed judges-to-cases ratio on the judicial system manifests in an overwhelming backlog of pending cases alongside an ongoing influx of new ones. To tackle this issue and expedite the judicial process, the proposition of an automated system capable of suggesting case outcomes based on factual evidence and precedent from past cases gains significance. This research paper centres on developing a graph neural network-based model to address the Legal Judgment Prediction (LJP) problem, recognizing the intrinsic graph structure of judicial cases and making it a binary node classification problem. We explored various embeddings as model features, while nodes such as time nodes and judicial acts were added and pruned to evaluate the model's performance. The study is done while considering the ethical dimension of fairness in these predictions, considering gender and name biases. A link prediction task is also conducted to assess the model's proficiency in anticipating connections between two specified nodes. By harnessing the capabilities of graph neural networks and incorporating fairness analyses, this research aims to contribute insights towards streamlining the adjudication process, enhancing judicial efficiency, and fostering a more equitable legal landscape, ultimately alleviating the strain imposed by mounting case backlogs. Our best-performing model with XLNet pre-trained embeddings as its features gives the macro F1 score of 75\% for the LJP task. For link prediction, the same set of features is the best performing giving ROC of more than 80\%.

\end{abstract}

\section{Introduction}
Many cases are pending in the Indian judiciary, and the courts face a meagre judge-to-case ratio\footnote{\url{https://www.tribuneindia.com/news/archive/comment/backlog-of-cases-crippling-judiciary-776503}}, which requires a fair, reliable, and automated system to predict the verdict of a case. One such task developed recently is the Legal Judgement Prediction (LJP). It aims to predict and suggest the judgement decisions of a court case based on facts and aid in how judgements have passed in previous years. With the access and development of large legal datasets, conducting studies on these tasks becomes imperative. In our work, we model the Indian judiciary using graphs due to its inherent inter-connected structure of cases and laws.

We also aim to analyse the effect of time and acts on the outcome of the cases. To ensure a verdict is fair, we check our model for bias to discover how fair the decisions are. Extending the task to link prediction, we check how well the model understands the relationship between two cases by predicting whether an edge exists between their corresponding graph nodes.
In the paper, we answer the following four research questions:\\
\textbf{RQ1:} How does a real-world setting like graph neural networks perform on the task of LJP?\\
\textbf{RQ2:} How does the model behave when we prune or add time and act nodes?\\
\textbf{RQ3:} How does the model perform when trained temporally, i.e. trained till a particular period and how well can the model predict an edge between two given nodes?\\
\textbf{RQ4:} How fair are the decisions in the task of LJP?\\

Following is the summary of the research contributions done in this paper:
\begin{enumerate}
\item We employ a graph neural network (GNN) under different embedding settings by adding and removing two different node characteristics, i.e. time and acts.
\item A link prediction task to observe how well the model can predict an edge between two nodes representing a particular case citing another case.
\item A set of temporal experiments to see the effect of time on the training of the model.
\item For fairness, we check how biased the model is while making predictions.
    
\end{enumerate}

\section{Literature Review}

There has been a great deal of research on the text in the legal domain and various tasks have been suggested, such as prior case retrieval \cite{jackson2003information}, crime classification \cite{wang2019hierarchical}, and judgment prediction \cite{zhong2020iteratively}.

For the LJP challenge, various strategies and corpora have been introduced. In order to get around BERT's input token count restriction for the LJP problem, \cite{chalkidis2019neural} presented a hierarchical variant of BERT \cite{devlin2018bert}. Using datasets from the Chinese AI and Law Challenge (CAIL2018), \cite{yang2019legal} deployed a Multi-Perspective Bi-Feedback Network to forecast the corresponding legal accusations, offences, and periods of punishment. On three Chinese datasets (CJO, PKU, and CAIL), \cite{zhong2018legal} used topological multi-task learning on a directed acyclic network to predict charges, including theft, traffic violation, and deliberate homicide.

To predict the charges on a dataset of Criminal Law of the People's Republic of China, \cite{luo2017learning} suggested an attention-based model given the case's facts and the relevant articles. Similarly, in a few-shot configuration, \cite{hu2018few} implemented an attribute-attentive model based on the case's facts. Using a legal reading comprehension technique on a Chinese dataset, \cite{long2019automatic} predicted the case's outcome. Given the facts and charges on a dataset created from documents of the Supreme People's Court of China, \cite{chen2019charge} used a deep gating network to predict prison terms. \cite{aletras2016predicting} employed a linear support vector machine (SVM) to predict violations based on the facts of cases from the European Court of Human Rights. \cite{sulea2017predicting} implemented SVM in the LJP task on cases from the French Supreme Court. \cite{katz2017general} proposed a random forest model to forecast the judges' ``Reverse'', ``Affirm'', and ``Other'' judgments in the US Supreme Court.

In their research, \cite{gan2021judgment} proposed a method for representing legal knowledge using logic rules in a co-attention network, which improves interpretability and logical reasoning. They demonstrate the effectiveness of their approach through comprehensive experiments conducted on a civil loan scenario. Similarly, \cite{ma2021legal} utilizes a real courtroom dataset to predict legal judgments. Using multi-task learning, they extensively analyze multi-role dialogues, including plaintiff's claims and court debate data, to understand facts and discriminate claims for final judgments. The works of \cite{yue2021neurjudge} introduce NeurJudge, a framework for predicting legal judgments that consider crime circumstances. They leverage intermediate subtask results to identify and utilize different circumstances for predicting other subtasks.

Another approach by \cite{ma2020judgment} employs LSTM \cite{hochreiter1997long} to predict legal judgments by comprehensively understanding case inputs, court debates, and multi-role dialogues. They also utilize multi-task learning to discriminate claims and reach final judgments. \cite{huang2021dependency} propose a unified text-to-text Transformer for LJP, where the auto-regressive decoder's dependencies among sub-tasks can be naturally established. They highlight the advantage of establishing dependencies among sub-tasks.

Furthermore, \cite{xu2020distinguish} uses a graph neural network to differentiate confusing charges. They leverage a novel attention mechanism to automatically learn subtle differences between law articles and extract effective discriminative features from fact descriptions. \cite{dong2021legal} employ a graph neural network (GNN) to address the LJP problem as a node classification task on a global consistency graph derived from the training set. They utilize a masked transformer network for case-aware node representations and leverage relational learning for local consistency through neighbours' label distribution. Variational expectation minimization optimizes both the node encoder and classifier.

\cite{li2019mann} introduce MANN, a multichannel attentive neural network model for the integrated LJP task. MANN learns from previous judgment documents and utilizes attention-based neural networks to capture latent feature representations focused on case facts, defendant persona, and relevant law articles. A two-tier structure empowers attentive sequence encoders to hierarchically model semantic interactions at word and sentence levels in the case description.

In their work, \cite{kapoor2022hldc} introduce the Hindi Legal Documents Corpus (HLDC) consisting of over 900K legal documents in Hindi. They also propose a Multi-Task Learning (MTL) based model incorporating summarization as an auxiliary task alongside the primary task of bail prediction. \cite{malik2021ildc} present ILDC, a vast corpus containing 35k Indian Supreme Court cases annotated with original court decisions. They explore various baseline models for case predictions and propose a hierarchical occlusion-based model to enhance explainability. \cite{gan2022exploiting} suggest a moco-based supervised contrastive learning approach to acquire distinguishable representations and determine optimal positive example pairs for all three LJP subtasks. They also enhance fact description representation by incorporating pre-trained numeracy models to utilize crime amounts for predicting penalty terms.

\section{Experiments}

For our experiments, we use graph neural network architecture (GraphSAGE \cite{hamilton2017inductive}) for node classification where nodes are cases, acts and time nodes. Text embeddings were used as node features and time \& act nodes are referred to as characteristic nodes as they are hypothesised to have an impact on the graphical model's performance. Further, to analyse the impact of time on verdicts, we divide the dataset into train and test based on the year up to which we want to train the model. For our experiments, we use the ILDC dataset curated by \cite{malik2021ildc}, in which they provide preprocessed cases for the task with their corresponding binary labels. In each case brought before the Supreme Court of India (SCI), the judge or panel determines whether the assertions made by the appellant/petitioner against the respondent should be deemed as "accepted" or "rejected" and accordingly, labels are assigned in the dataset. The dataset already has split examples into train, test and development sets(5082/1517/994). We use the train and development splits as the train split and the test split as itself.\\
As we employed the graph neural network for our experiments, we had to increase the size of the dataset because the dataset 
\cite{malik2021ildc} does not provide the cases that were cited by the cases. Using the ikanoon API\footnote{\url{https://api.indiankanoon.org/}}, we extracted 24,907 additional cases and added them as nodes to the graph network to complete the citation network. This achieved the semi-supervised setting, enabling message passing between nodes and learning from other cases.

\input{tables/ildc_year_stat}

\subsection{Different embeddings}

\begin{itemize}
    \item \textbf{Random} We initialize random embeddings as node features to train the model.
    \item \textbf{XLNet} We initialize XLNet embeddings as node features to train the model.
    \item \textbf{XLNet Pretrained} We take the previously trained XLNet model on the task of judgment prediction from \cite{malik2021ildc} and extract embeddings for the nodes to train the model.
    \item \textbf{hierarchical} For the train split in the dataset, we initialize embeddings from XLNet pre-trained on the judgment prediction, and for the test split, we initialize embeddings from XLNet (not pre-trained).
\end{itemize}

\subsection{Edge Types}
Another set of experiments included the edge type between two nodes in the graph.

\begin{itemize}
    \item \textbf{Directed:} Given a case, an edge is directed from it to all the cases it cites, enabling message-passing from that case node to the cited case node.
    \item \textbf{Rev-Directed:} Given a case, an edge is directed to it from all the cases it cites, enabling message-passing to the case node from the cited case node. It is the most practical use case as the message has to be passed between various nodes to make the network aware of the legal knowledge and decisions.
    \item \textbf{Undirected:} Given a case, an undirected edge is present between it and the cases it cites, enabling message-passing from that case node to the cited case node and vice versa.
\end{itemize}

\input{tables/simple_jp}
\begin{figure*}[!ht]
    \centering
    \resizebox{\textwidth}{12cm}{%
    \includegraphics{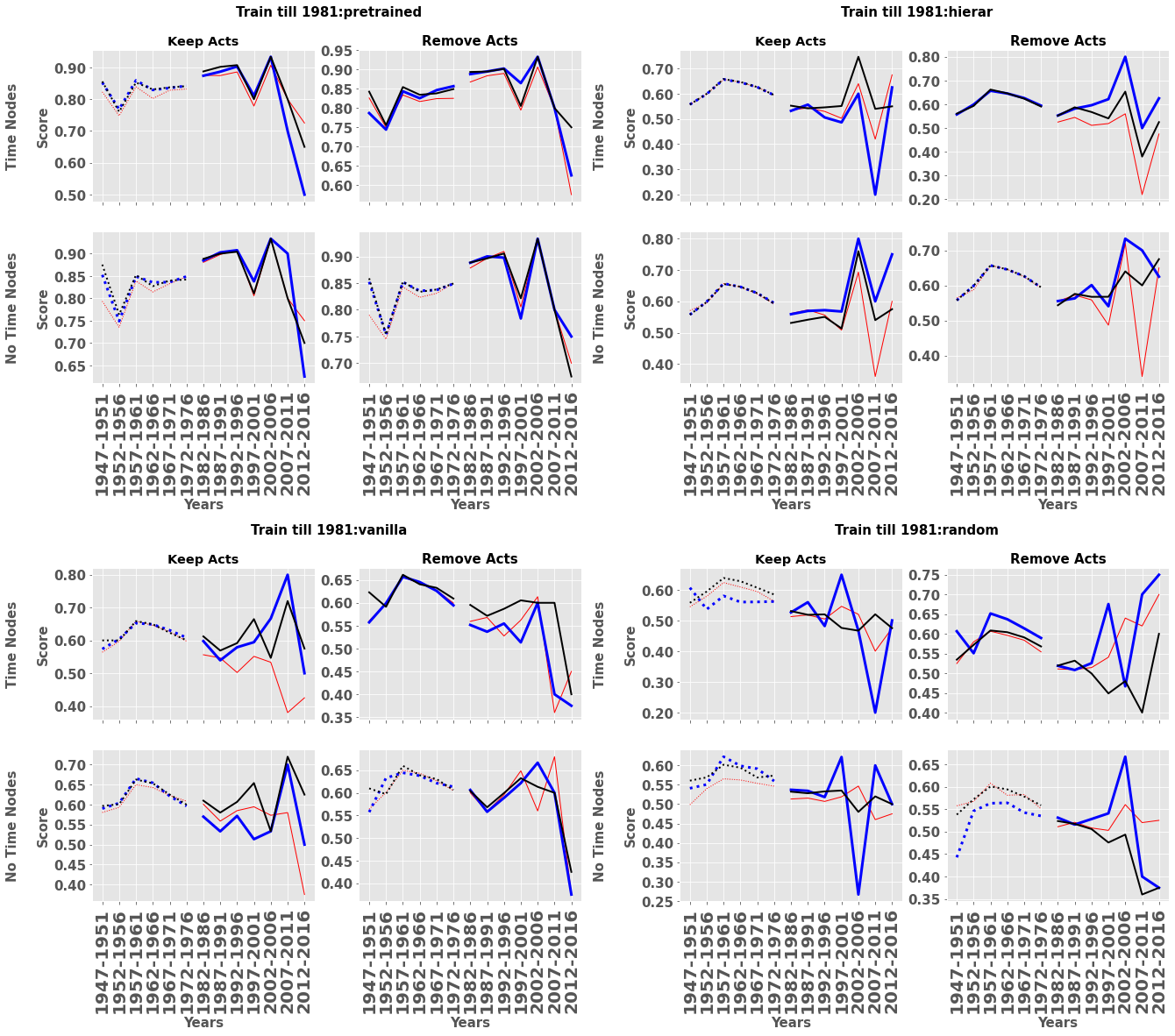}%
    }
    \caption{Judgment Prediction Task: Results of temporal training on different embeddings. The thickest line represents an undirected edge, the less thick line represents the rev\_edges setting, and the thinnest line represents the no\_change setting. The broken line represents that training is done in a reversed manner.}
    \label{jp_year1981}
\end{figure*}

\subsection{Time}

The dataset contains cases from the year 1956 to the year 2021. To study the impact of time on these verdicts, we added new nodes, called time nodes, in the graph with their connection to the corresponding case in that particular year. The node features of these time nodes are randomly initialised embeddings.

\subsection{Acts}
Cases come with supporting arguments referencing acts from the Indian constitution to make an argument better. We experiment with the retention and removal of those act nodes in the graph model to observe how the model prediction changes based on underlying act nodes with their connection to the case.

\subsection{Simple Training}
A graph network is trained on the dataset as specified by the train and test splits \cite{malik2021ildc}. Experimentation involved training with and without characteristic nodes. We recorded the macro precision, recall and F1 scores.

\subsection{Temporal Training}

To study the temporal aspect of how a verdict is made based on cases and acts present in that particular split and how the law has changed over time. The model is trained on cases for a particular number of years and tested on the rest.


\begin{figure*}[!ht]
\begin{subfigure}[t]{0.5\textwidth}
\includegraphics[width=0.98\textwidth]{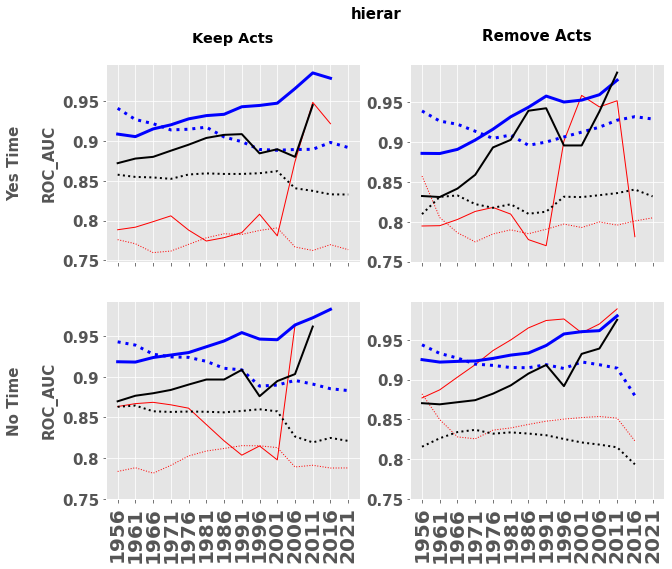}    
\end{subfigure}
\hfill
\begin{subfigure}[t]{0.5\textwidth}
\includegraphics[width=0.98\textwidth]{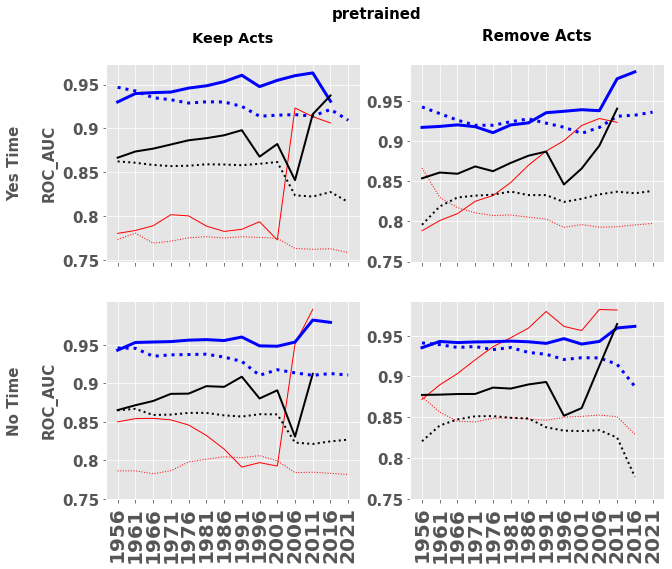} 
\end{subfigure}

\begin{subfigure}[t]{0.5\textwidth}
\includegraphics[width=0.98\textwidth]{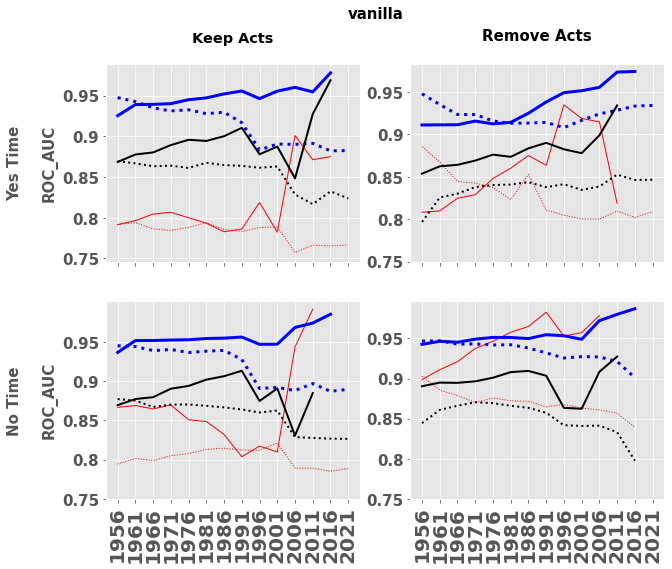} 
\end{subfigure}
\begin{subfigure}[t]{0.5\textwidth}
\includegraphics[width=0.98\textwidth]{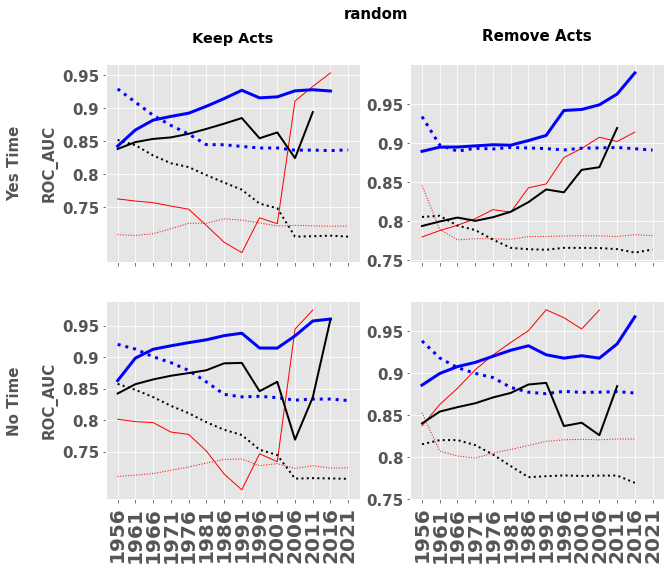}
\end{subfigure}
\caption{Link Prediction Task: Results of temporal training on different embeddings. The thickest line represents an undirected edge, the less thick line represents the rev\_edges setting, and the thinnest line represents the no\_change setting. The broken line represents that training is done in a reversed manner.}
\label{link_prediction}
\end{figure*}

\subsubsection{Training direction}
\begin{itemize}
    \item \textbf{Forward} We train the model for a particular number of years and then predict future case verdicts. For example, we train the model till 2001 and then predict the verdicts from 2002 to 2021 to check how verdicts present till a particular year affect future cases. We have a range from 1956 to 2021. We tend to predict cases in 5 years from the year of training till the range, i.e. if the model is trained till 1956, predictions are made on cases from 1957-1961, 1962-1966, 1967-1971 and so on till 2021.
    
    \item \textbf{Reverse} We train the model for a particular number of years and then predict past case verdicts. For example, we train the model from 2001 till 2021 and then predict the verdicts from 1956 to 2000 to check how new laws can comprehend past cases and make good predictions. We tend to predict cases in 5 years from the year of training till the range, i.e. if the model is trained till 2001, predictions are made on years 1996-2000, 1991-1995, 1976-1990 and so on till 1956.
\end{itemize}

\subsection{Link Prediction}
Link prediction is predicting if any edge exists between two nodes. We experimented with the task of link prediction based on the temporal aspect of the cases. We used the complete dataset of 24,907 cases from the early 1800s to 2021, divided nodes into train and test, and ran the above experiments to observe how well the link prediction works in the given settings. We plotted the ROC curve Figure \ref{link_prediction} of the result.

\begin{figure*}[!ht]
    \centering
    \resizebox{\textwidth}{!}{%
    \includegraphics{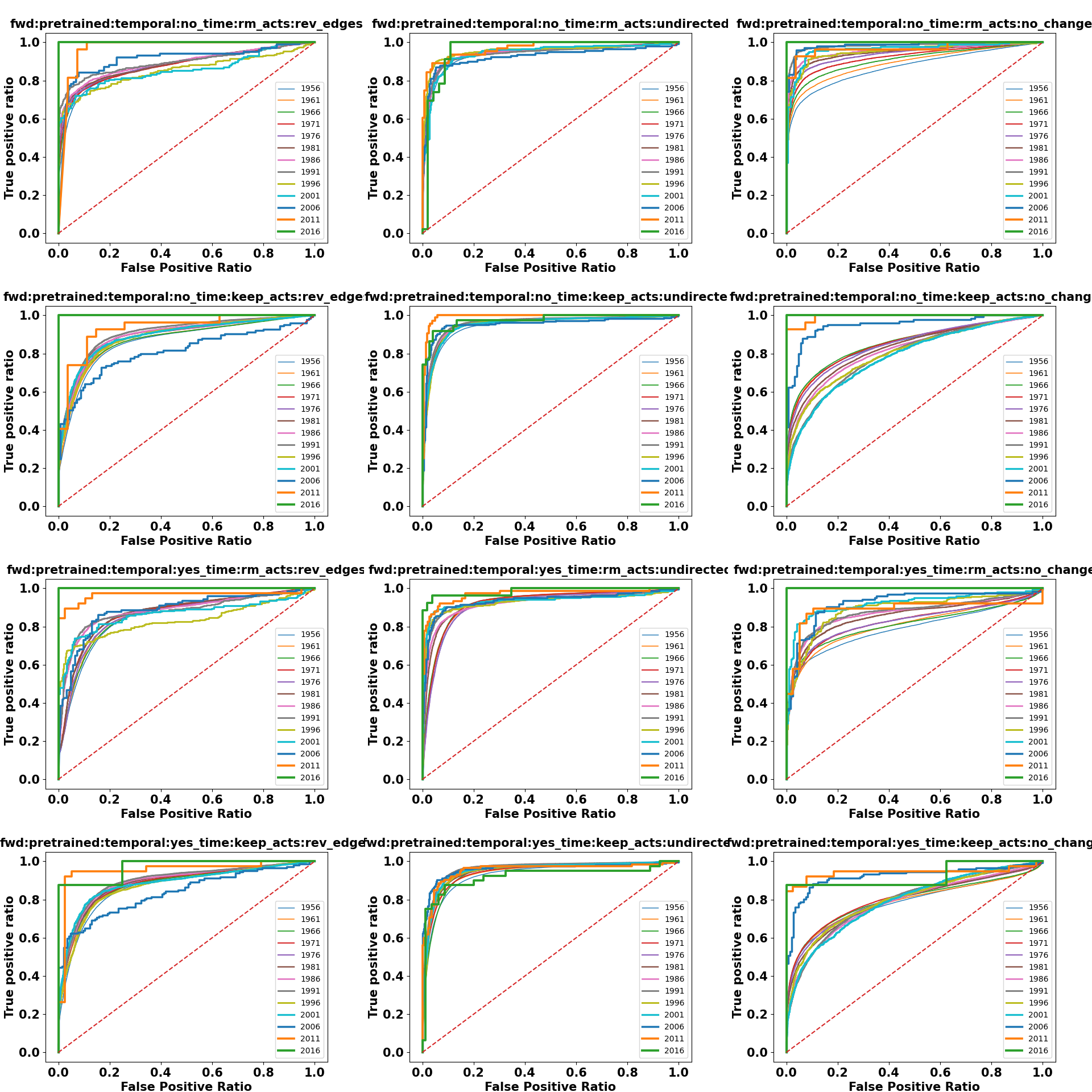}%
    }
    \caption{ROC curves for link prediction task using pre-trained embeddings. Line thickness increases from the earliest to the latest year in the dataset.}
    \label{lp_roc}
\end{figure*}

\subsection{Redaction}

We redacted gender-biased terms, in our case pronouns (he, him, his, her, she) and name tokens with \textit{[gender]} and \textit{[REDACTED]} tokens and fine-tuned the XLNet and graph models again on the above-given settings for link prediction. We used Indian legal NER\footnote{\url{https://huggingface.co/opennyaiorg/en\_legal\_ner\_trf}} to extract name entities.

\section{Results and Discussion}

\subsection{Simple training}

In Table \ref{simple_jp}, we can observe that the difference in the observations is mainly due to the different features i.e. the text embeddings used. Pretrained embeddings were the top features for the model, followed by vanilla XLNet embeddings, hierarchical and random. 
The type of edges between nodes has minimal effect. Moreover, adding and removing the time nodes and acts did not significantly change the model's performance. We can only observe a minute change in the model's performance when we add time nodes.

\subsection{Temporal Training}

\begin{figure*}[!ht]
    \centering
    \resizebox{\textwidth}{!}{%
    \includegraphics{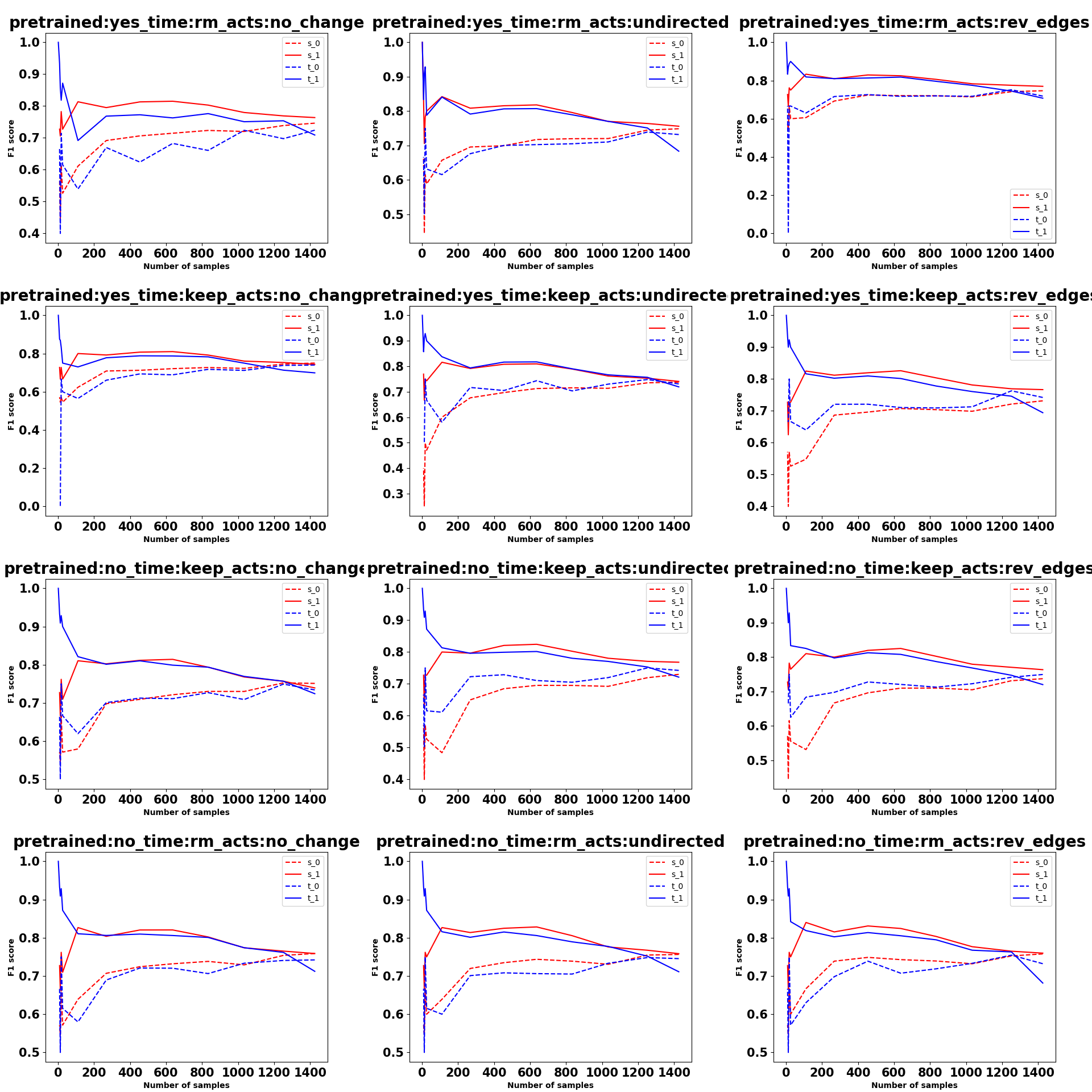}%
    }
    \caption{F1 scores of binary labels of the GNN model with pre-trained features calculated on different numbers of samples distributed according to the year as present in Table \ref{ildc_year_stat}. The dotted line represents label 0 and s\_0 represents the performance of label 0 training in the simple setting and continuous line represents label 1 and s\_1 represents the performance of label 1 while training in simple setting. Line t\_0 represents the performance of label 0 while training temporally. Line t\_1 represents the performance of label 1 while training temporally. The X-axis represents the number of training samples presented according to Table \ref{ildc_year_stat} and the Y-axis represents the F1 score. }
    \label{label_pt}
\end{figure*}

The trend for all the graphs except for the model with pre-trained embeddings is the same. Figure \ref{jp_year1981} shows that in the forward training direction, as we keep training the graph on consecutive years, the F1 score of the model keeps increasing as the data increases, resulting in a non-negative slope from a negative slope. Training in the reverse direction gives us a negative slope over the years, indicating that trained on future predictions, the model performs poorly on past verdicts when only the future data is used to train the model.
The model with pre-trained features gets roughly the same F1 score for both classes on the test data; see Figure \ref{label_pt}. Comparison is also made based on the intersection of test samples in the simple and the created temporal datasets. However, when we only take the test samples from the temporal dataset, we can see that the model has an F1 score in the range of around 80\%-85\% across all years, and we can see a drop after 2011. This observation also includes samples used to pretrain XLNet to get text embeddings. To confirm that the model is not learning irrelevant patterns, as some instances are used to train XLNet, as mentioned, and that the F1 score is not based on pre-trained embeddings, we shuffled the labels by 50\% \cite{howard2020fastai} and trained the model. In every case, we got an F1 score lower than 50\%, confirming that the close F1 scores by the model in temporal and simple settings are not random, and the model is learning the patterns.
For the vanilla embeddings, we can see in Figure \ref{label_van} that it performs better than simple training in some scenarios, like when time nodes are not added to the model.
After 1981, we can see the kink changing its direction in the graphs from and after 2001-2006; this is observed in the models with features except for pre-trained embeddings.

\subsection{Link Prediction}

For the link prediction task, in forward training, we can see in Figure \ref{link_prediction} when the edges are undirected, the model can predict it more efficiently, followed by rev\_edges and no\_change. The embeddings have significantly less effect in the forward training direction.

Figure \ref{lp_roc} shows ROC curves broken down to years for each pre-trained embedding and each setting. More are present in the appendix. The area under the curve (AUC) for pre-trained is the most, followed by hierarchical, vanilla and random. 

The addition of time and act nodes decreases the AUC of the curve, which means the model is not very efficient in predicting an edge between a case and time and\textbackslash or act nodes.

Reverse training also exhibited the same observations except for the part where the AUC was lower than forward training, specifically for the reverse training setting.

\subsection{Redaction}

Training our model with redacted tokens gives almost the same output for all the models compared to the unredacted dataset, hinting at little to no bias in the judgments. 

\section{Conclusion}
By this study, we conclude that the time nodes have a negative impact on the performance of the model, as we can observe in Figure  \ref{label_pt}, \ref{label_van}, \ref{label_hierar} and \ref{label_random} where the difference F1 score of both the classes is more significant with the time nodes. The opposite is observed in the case of acts; when we have the act nodes, the difference between F1 scores of both the classes reduces, which implies the best setting is with no time nodes and keeping act nodes in the model. Secondly, embeddings have the least significant impact on the task of link prediction, and it mainly predicts the presence of an edge between two nodes. Predicting the edge's direction is challenging for the model as undirected edges are predicted better. Lastly,  we found no significant change in the model's performance and classes with the redacted tokens.   

\section{Limitations}

As per the limitations, the explainability of judgments was out of the scope of the paper. We limit ourselves to existing graph models for our study. We mainly focus on a more profound analysis of the LJP task using graph neural networks, so we are on par with the best-performing model proposed in \cite{malik2021ildc}.

\section{Ethics}

Our study aims to advance research and automation in the legal domain,  focusing on the Indian legal system. We are committed to making the extended dataset and resources we use publicly available, ensuring accessibility for all. Given the substantial number of pending cases in lower courts, our efforts are directed towards enhancing the legal system, which stands to benefit millions of people. Our work aligns with previous efforts in legal NLP, such as creating legal corpora and predicting legal judgments.

Nevertheless, we acknowledge the potential risks associated with developing AI systems based on legal corpora, which could adversely affect individuals and society. To address this concern, we took proactive measures to identify and mitigate biases in the corpus. One of these measures includes anonymizing entities, such as names and gender, in the dataset. This is particularly important as previous research \cite{malik2021ildc} and \cite{kapoor2022hldc} have shown biases in legal datasets.

Additionally, we explored the task of Link Prediction in the Indian legal domain, which holds significance for developing recommendation systems in the legal field. This area is relatively novel in NLP research and remains nascent in India. Consequently, further research and investigations are essential, especially concerning potential biases and societal impacts.

\section{Acknowledgements}
We would like to acknowledge iHub Anubhuti IIIT Delhi for funding our research which enabled us to get the required resources.

%
%
%
\bibliographystyle{splncs04}
\bibliography{bda}

\clearpage
\appendix

\section{Appendix}
\begin{figure*}[!ht]
    \centering
    \resizebox{\textwidth}{!}{%
    \includegraphics{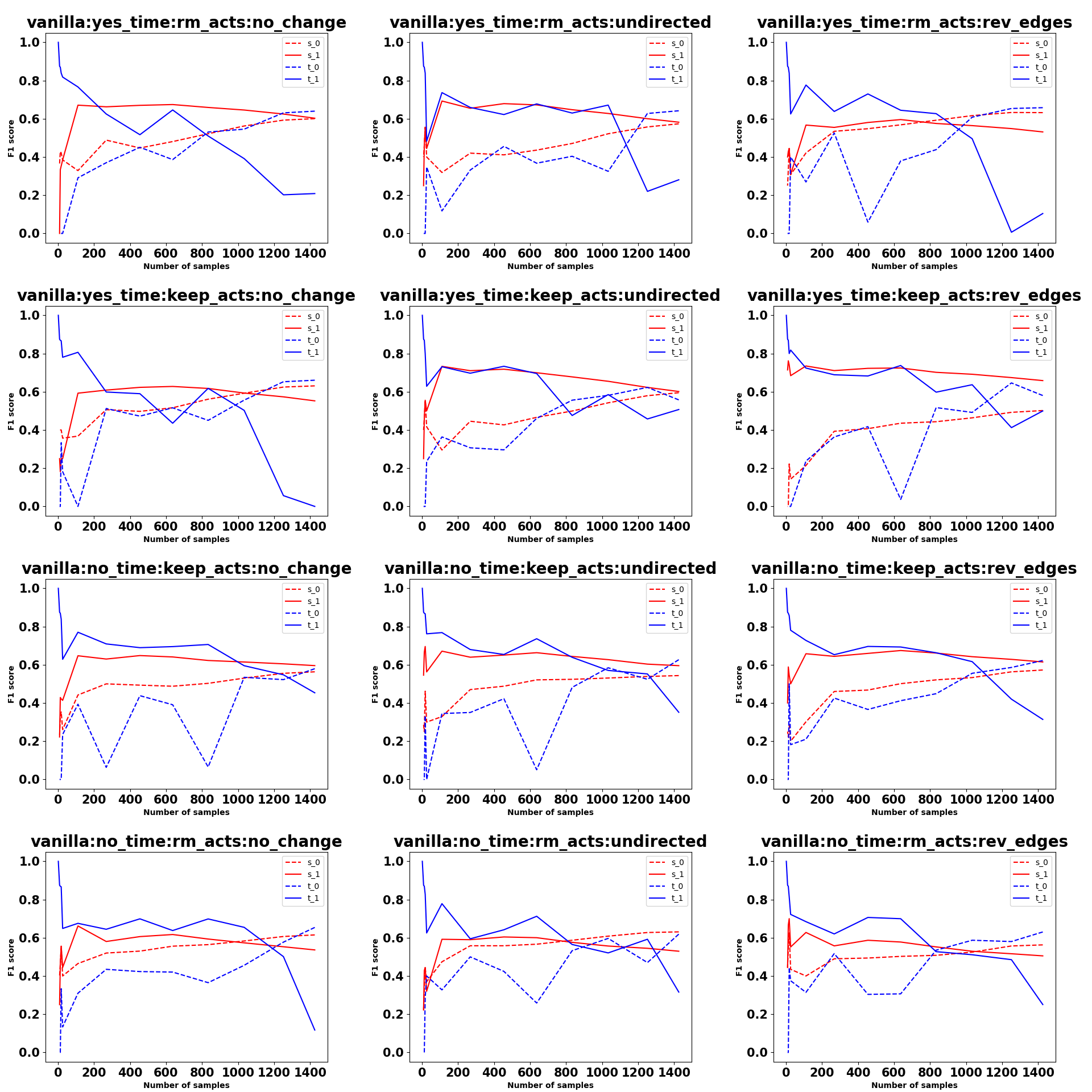}%
    }
    \caption{F1 score of binary labels of GNN model with vanilla features calculated on different number of samples distributed according to the year as present in Table \ref{ildc_year_stat}. Dotted line represents label 0 and s\_0 represents performance of label 0 while training in simple setting and continuous line represents label 1 and s\_1 represents performance of label 1 while training in simple setting. Line t\_0 represents performance of label 0 while training temporally. Line t\_1 represents performance of label 1 while training temporally. X-axis is number of training samples presented according to Table \ref{ildc_year_stat} and Y-axis represents F1 score.}
    \label{label_van}
\end{figure*}

\begin{figure*}
    \centering
    \resizebox{\textwidth}{!}{%
    \includegraphics{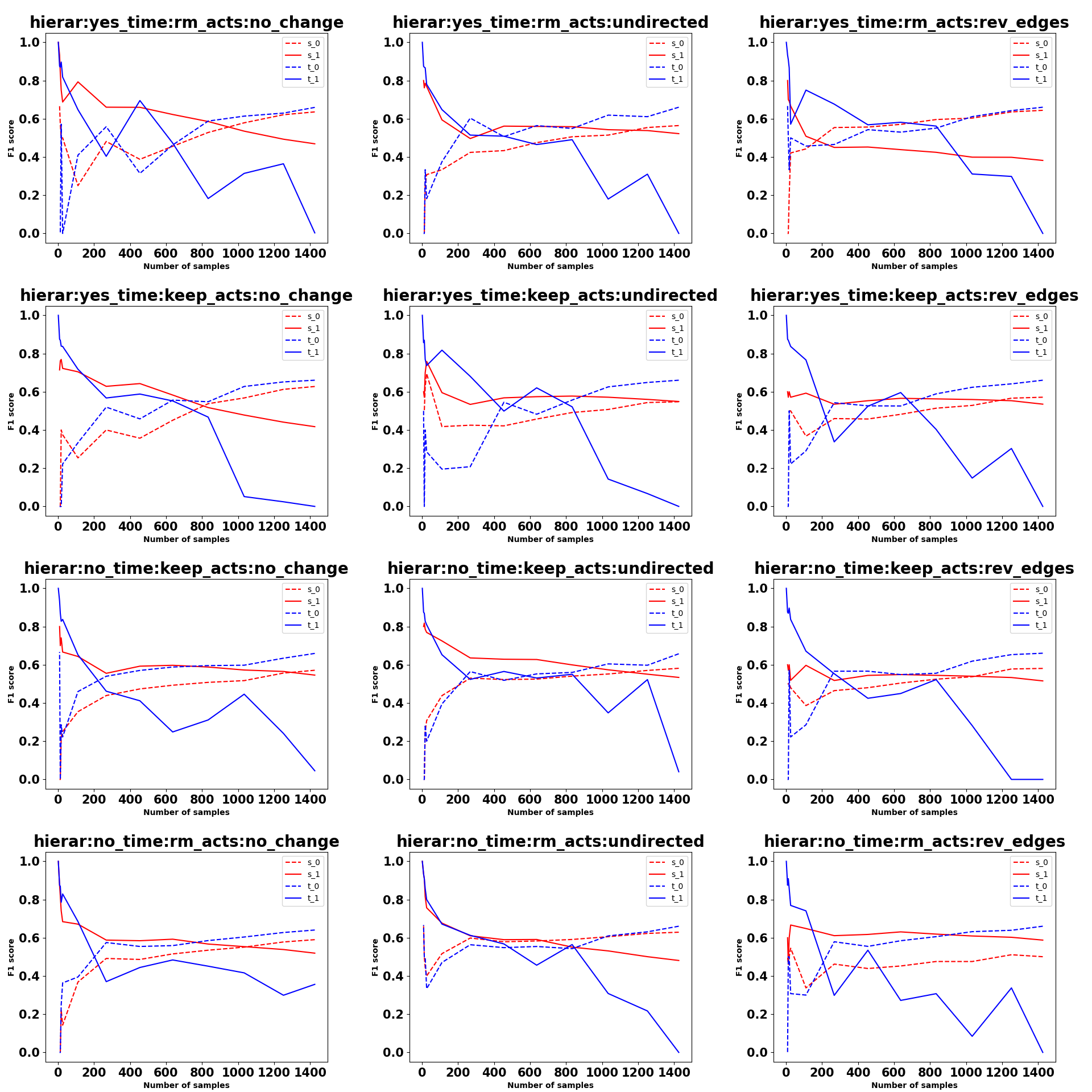}%
    }
    \caption{F1 score of binary labels of GNN model with hierar features calculated on different number of samples distributed according to the year as present in Table \ref{ildc_year_stat}. Dotted line represents label 0 and s\_0 represents performance of label 0 while training in simple setting and continuous line represents label 1 and s\_1 represents performance of label 1 while training in simple setting. Line t\_0 represents performance of label 0 while training temporally. Line t\_1 represents performance of label 1 while training temporally. X-axis is number of training samples presented according to Table \ref{ildc_year_stat} and Y-axis represents F1 score.}
    \label{label_hierar}
\end{figure*}

\begin{figure*}
    \centering
    \resizebox{\textwidth}{!}{%
    \includegraphics{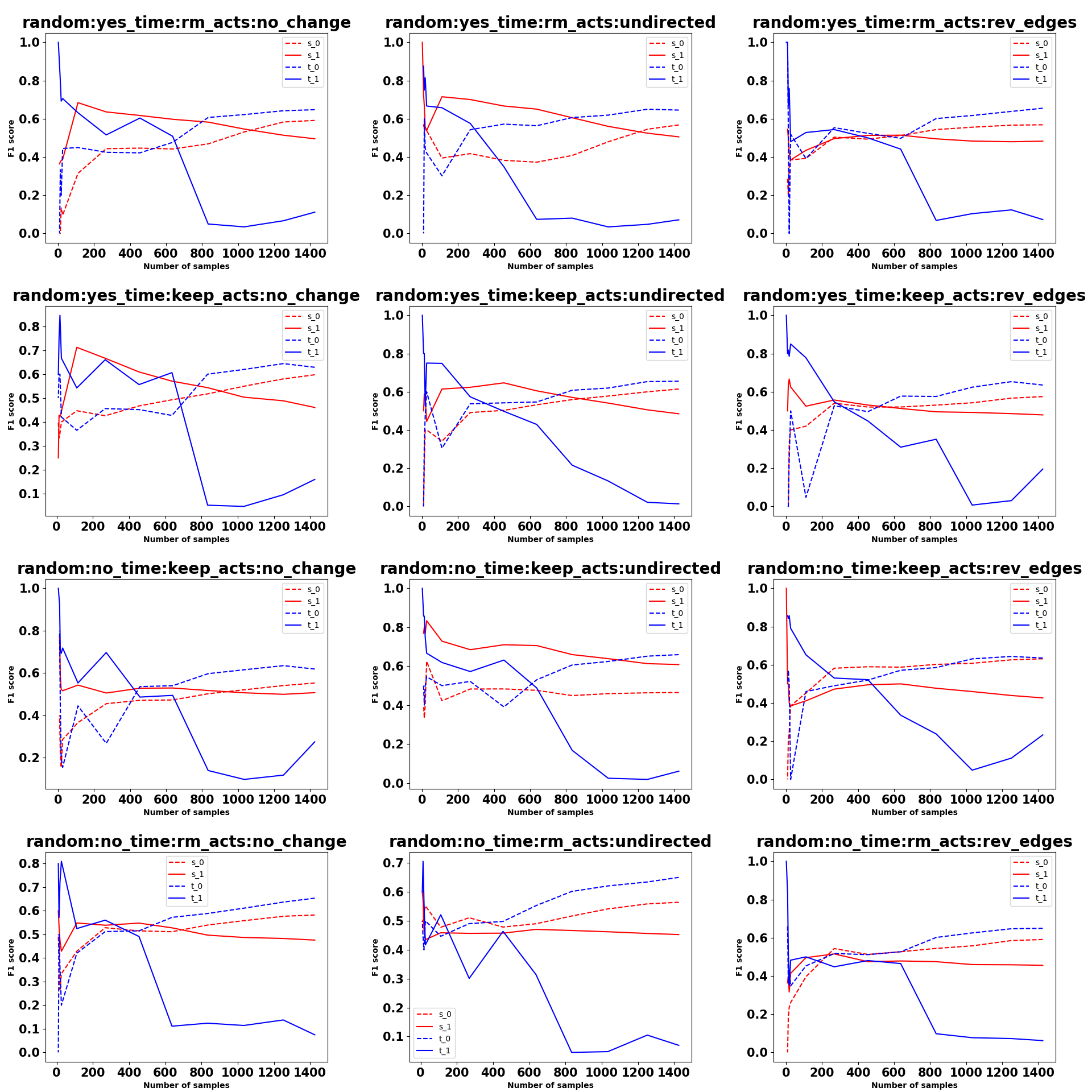}%
    }
    \caption{F1 score of binary labels of GNN model with random features calculated on different number of samples distributed according to the year as present in Table \ref{ildc_year_stat}. Dotted line represents label 0 and s\_0 represents performance of label 0 while training in simple setting and continuous line represents label 1 and s\_1 represents performance of label 1 while training in simple setting. Line t\_0 represents performance of label 0 while training temporally. Line t\_1 represents performance of label 1 while training temporally. X-axis is number of training samples presented according to Table \ref{ildc_year_stat} and Y-axis represents F1 score.}
    \label{label_random}
\end{figure*}


\begin{figure*}[!ht]
    \centering
    \resizebox{\textwidth}{!}{%
    \includegraphics{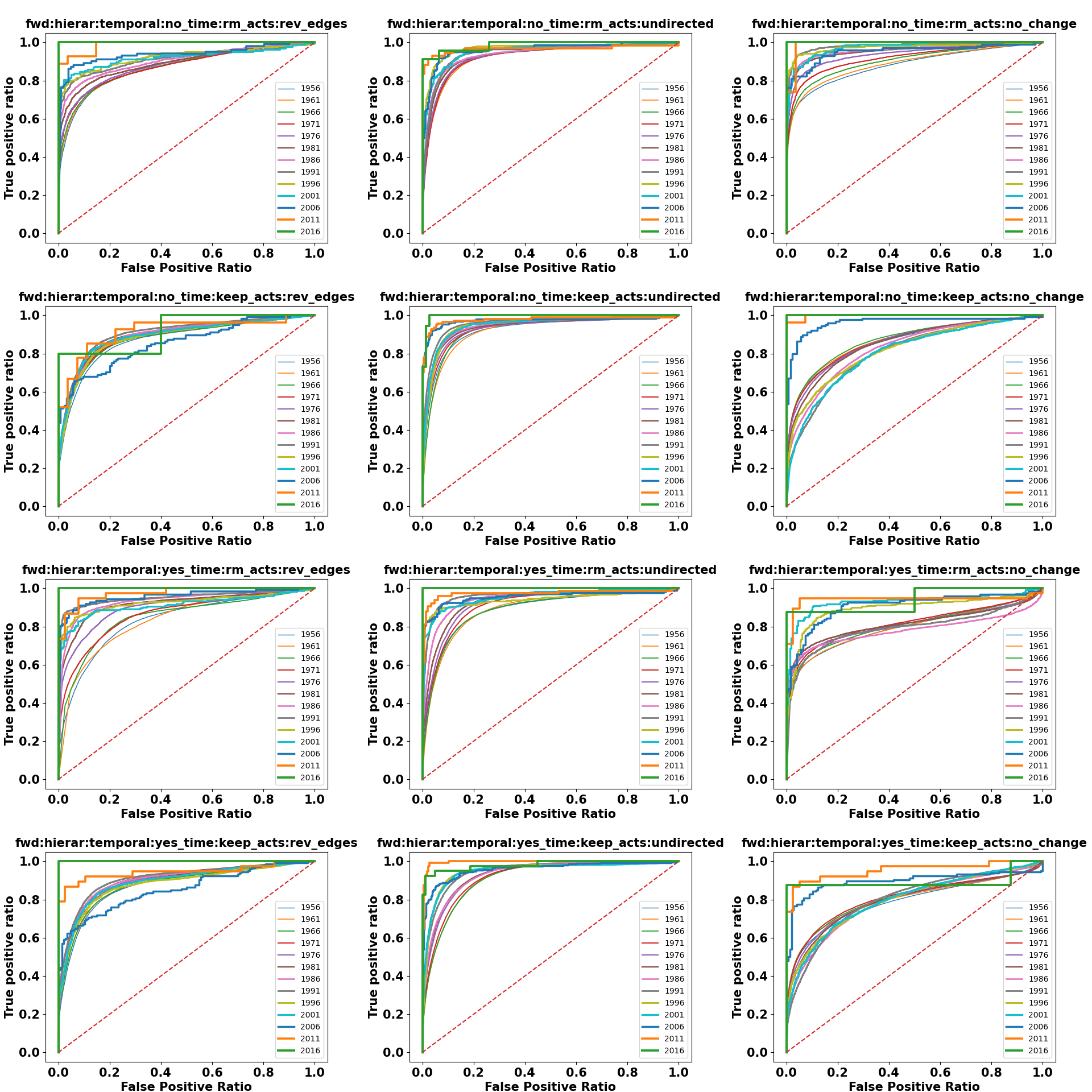}%
    }
    \caption{ROC curves across years for hierarchical embeddings trained in forward direction}
    \label{lp_roc_hierar_fwd}
\end{figure*}

\begin{figure*}[!ht]
    \centering
    \resizebox{\textwidth}{!}{%
    \includegraphics{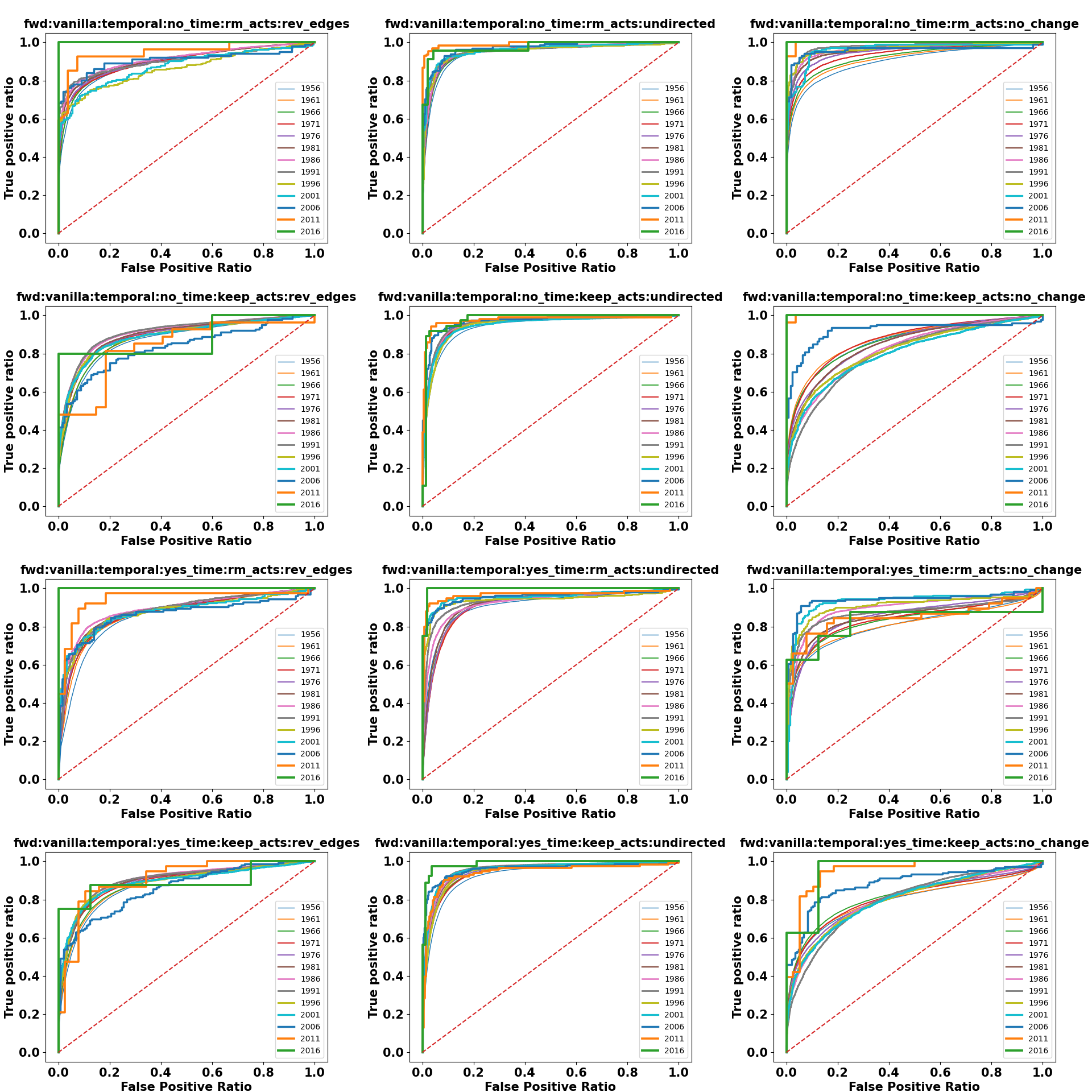}%
    }
    \caption{ROC curves across years for vanilla embeddings trained in forward direction}
    \label{lp_roc_vanilla_fwd}
\end{figure*}

\begin{figure*}[!ht]
    \centering
    \resizebox{\textwidth}{!}{%
    \includegraphics{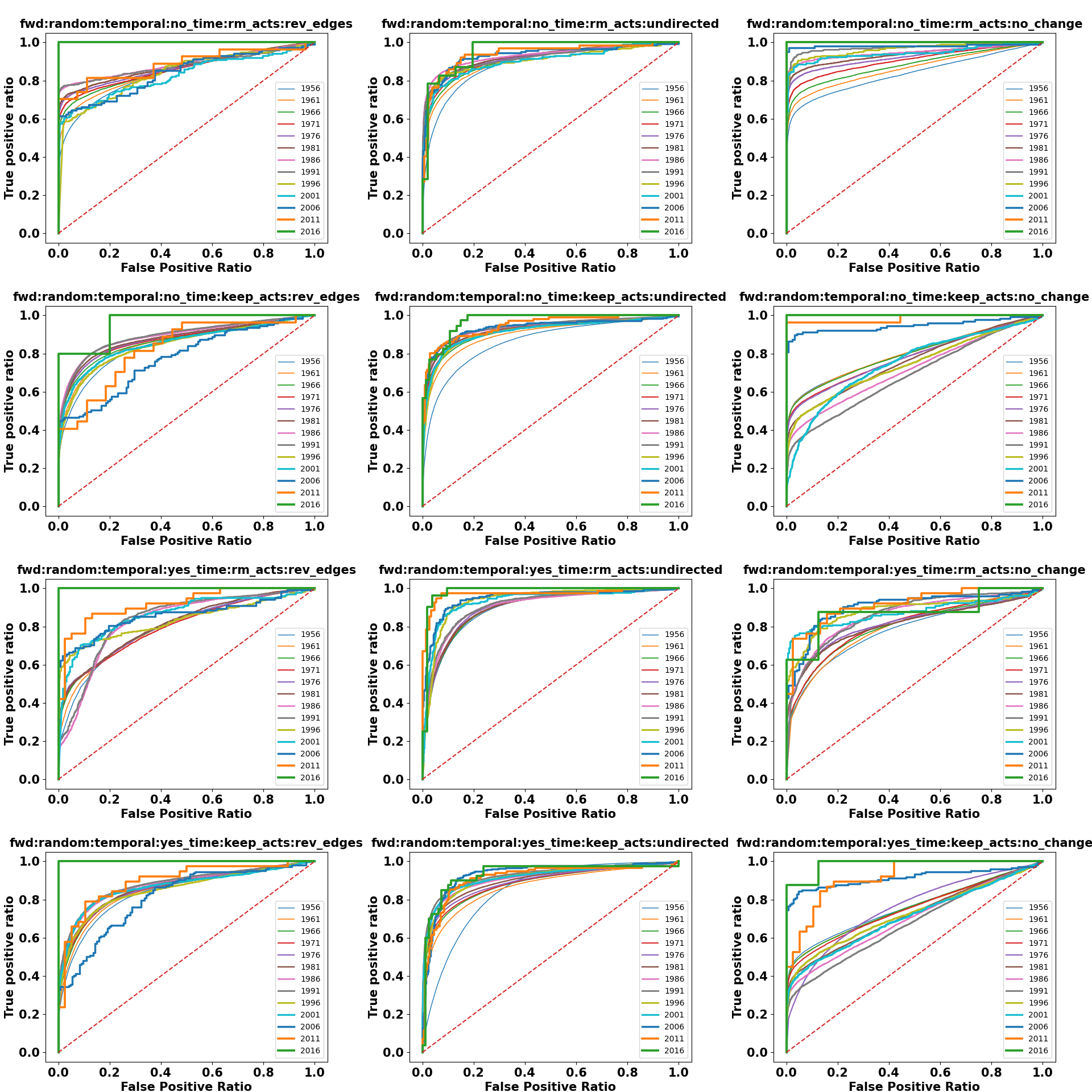}%
    }
    \caption{ROC curves across years for random embeddings trained in forward direction}
    \label{lp_roc_random_fwd}
\end{figure*}





\end{document}

%% file: tables/ildc_year_stat.tex
\begin{table}[!ht]
\centering
\begin{tabular}{|c|c|c|}
\hline
\textbf{Year range} & \textbf{\begin{tabular}[c]{@{}c@{}}Number of \\ Training \\ Examples\end{tabular}} & \textbf{\begin{tabular}[c]{@{}c@{}}Cumulative \\ Frequency\end{tabular}} \\ \hline
1956-1960           & 447                                                                                & 447                                                                      \\ \hline
1961-1965           & 955                                                                                & 1402                                                                     \\ \hline
1966-1970           & 714                                                                                & 2116                                                                     \\ \hline
1971-1975           & 728                                                                                & 2844                                                                     \\ \hline
1976-1980           & 622                                                                                & 3466                                                                     \\ \hline
1981-1985           & 515                                                                                & 3981                                                                     \\ \hline
1986-1990           & 547                                                                                & 4528                                                                     \\ \hline
1991-1995           & 366                                                                                & 4894                                                                     \\ \hline
1996-2000           & 24                                                                                 & 4918                                                                     \\ \hline
2001-2005           & 10                                                                                 & 4928                                                                     \\ \hline
2006-2010           & 5                                                                                  & 4933                                                                     \\ \hline
2011-2015           & 1                                                                                  & 4934                                                                     \\ \hline
2016-2020           & 1                                                                                  & 4935                                                                     \\ \hline
\end{tabular}
\caption{Number of training examples distributed year-wise with their cumulative frequency.}
\label{ildc_year_stat}
\end{table}

%% file: tables/simple_jp.tex
\begin{table*}[!ht]
\resizebox{\textwidth}{1.2cm}{%
\begin{tabular}{|l|llll|llll|llll|}
\hline
\multirow{3}{*}{\textbf{Embeddings}} & \multicolumn{4}{l|}{\textbf{no\_change}}                                                                                                       & \multicolumn{4}{l|}{\textbf{undirected}}                                                                                                       & \multicolumn{4}{l|}{\textbf{rev\_edges}}                                                                                                       \\ \cline{2-13} 
                                     & \multicolumn{2}{l|}{\textbf{rm\_acts}}                                           & \multicolumn{2}{l|}{\textbf{keep\_acts}}                    & \multicolumn{2}{l|}{\textbf{rm\_acts}}                                           & \multicolumn{2}{l|}{\textbf{keep\_acts}}                    & \multicolumn{2}{l|}{\textbf{rm\_acts}}                                           & \multicolumn{2}{l|}{\textbf{keep\_acts}}                    \\ \cline{2-13} 
                                     & \multicolumn{1}{l|}{\textbf{yes\_time}} & \multicolumn{1}{l|}{\textbf{no\_time}} & \multicolumn{1}{l|}{\textbf{yes\_time}} & \textbf{no\_time} & \multicolumn{1}{l|}{\textbf{yes\_time}} & \multicolumn{1}{l|}{\textbf{no\_time}} & \multicolumn{1}{l|}{\textbf{yes\_time}} & \textbf{no\_time} & \multicolumn{1}{l|}{\textbf{yes\_time}} & \multicolumn{1}{l|}{\textbf{no\_time}} & \multicolumn{1}{l|}{\textbf{yes\_time}} & \textbf{no\_time} \\ \hline
\textbf{vanilla}                     & \multicolumn{1}{l|}{58.53}              & \multicolumn{1}{l|}{57.84}             & \multicolumn{1}{l|}{59.69}              & 57.45             & \multicolumn{1}{l|}{59.12}              & \multicolumn{1}{l|}{58.49}             & \multicolumn{1}{l|}{58.64}              & 58.97             & \multicolumn{1}{l|}{58.3}               & \multicolumn{1}{l|}{58.37}             & \multicolumn{1}{l|}{59.28}              & 59.62             \\ \hline
\textbf{pretrained}                  & \multicolumn{1}{l|}{\textbf{74.81}}              & \multicolumn{1}{l|}{\textbf{75.14}}             & \multicolumn{1}{l|}{\textbf{75.0}}                 & \textbf{74.98}             & \multicolumn{1}{l|}{\textbf{74.85}}              & \multicolumn{1}{l|}{\textbf{75.35}}             & \multicolumn{1}{l|}{\textbf{73.86}}              & \textbf{74.46}             & \multicolumn{1}{l|}{\textbf{75.35}}              & \multicolumn{1}{l|}{\textbf{75.3}}              & \multicolumn{1}{l|}{\textbf{74.49}}              & \textbf{74.22}             \\ \hline
\textbf{random}                      & \multicolumn{1}{l|}{53.80}               & \multicolumn{1}{l|}{52.75}                 & \multicolumn{1}{l|}{54.63}              & 53.21             & \multicolumn{1}{l|}{54.58}              & \multicolumn{1}{l|}{52.83}                 & \multicolumn{1}{l|}{55.72}              & 53.52             & \multicolumn{1}{l|}{52.10}               & \multicolumn{1}{l|}{53.24}                 & \multicolumn{1}{l|}{53.88}              & 53.05             \\ \hline
\textbf{hierar}                      & \multicolumn{1}{l|}{54.88}              & \multicolumn{1}{l|}{55.53}             & \multicolumn{1}{l|}{55.80}               & 55.80              & \multicolumn{1}{l|}{55.42}              & \multicolumn{1}{l|}{55.45}             & \multicolumn{1}{l|}{55.10}               & 54.9              & \multicolumn{1}{l|}{55.52}              & \multicolumn{1}{l|}{55.18}             & \multicolumn{1}{l|}{54.53}              & 54.49             \\ \hline
\end{tabular}%
}
\caption{Results from simple training. The table shows the different embeddings used with time and act settings. Pretrained embeddings were the best performing of all the embeddings used in the experiments. Acts and Time nodes were less significant in the model.}
\label{simple_jp}%
\end{table*}